%% file: main.tex
\documentclass[10pt,twocolumn,letterpaper]{article}

\usepackage{cvpr}
\usepackage{times}
\usepackage{epsfig}
\usepackage{graphicx}
\usepackage{amsmath}
\usepackage{amssymb}
\usepackage{enumitem}
\usepackage{booktabs}
\usepackage{url}
\usepackage{float}

\usepackage{dblfloatfix}

\usepackage{multirow}
\usepackage{color}
\usepackage{colortbl}
\definecolor{CuGray}{gray}{0.9}
\newcolumntype{g}{>{\columncolor{CuGray}}c}
\newcolumntype{z}{>{\columncolor{CuGray}}l}

\usepackage[flushmargin]{footmisc}
\input{DEF}

\newcommand{\figref}[1]{Fig.~\ref{fig:#1}}
\newcommand{\tabref}[1]{Table \ref{tab:#1}}
\usepackage{rotating}
\usepackage{comment}
\usepackage{microtype}
\newcommand{\beforepar}{\vspace{1mm}} 
\newcommand{\afterfig}{\vspace{-0.2cm}}

\usepackage[pagebackref=true,breaklinks=true,letterpaper=true,colorlinks,bookmarks=false]{hyperref}

\cvprfinalcopy 

\def\arxivversion

\def\cvprPaperID{3427} 

\ifcvprfinal\fi
\begin{document}


\title{\vspace{-1cm}Speech2Face: Learning the Face Behind a Voice}


\author{
Tae-Hyun Oh${}$\thanks{The three authors  contributed equally to this work.
\newline
Correspondence: \url{taehyun@csail.mit.edu}\newline Supplementary material (SM): \url{https://speech2face.github.io}
}${}^{*\dag}$\qquad
Tali Dekel${}^{*}$\qquad
Changil Kim${}^{*\dag}$ \qquad
Inbar Mosseri${}$ \\[1mm]
William T. Freeman$^{\dag}$ \qquad\qquad
Michael Rubinstein${}$ \qquad\qquad
Wojciech Matusik$^{\dag}$
\\[4mm]
$^{\dag}$MIT CSAIL
\vspace{-2mm}
}

\maketitle


\input{sec0_abstract.tex}
\input{sec1_introduction.tex}
\input{teaser.tex}
\ifdefined\arxivversion
    \input{appendix.tex}

\fi
\input{sec2_background.tex}

\input{sec3_implementation.tex}

\input{sec4_results.tex}

\input{sec5_conculsion.tex}

\vspace{2mm}
\paragraph{Acknowledgment}
The authors would like to thank Suwon Shon, James Glass, Forrester Cole and Dilip Krishnan for helpful discussion. T.-H. Oh and C. Kim were supported by QCRI-CSAIL Computer Science Research Program at MIT. 
%

{\small
\bibliographystyle{ieee}
\bibliography{refs}
}

\end{document}

%% file: DEF.tex
\newcommand{\be}{\begin{eqnarray}}
\newcommand{\ee}{\end{eqnarray}}
\newcommand{\bee}{\begin{eqnarray*}}
\newcommand{\eee}{\end{eqnarray*}}

\newcommand{\matrixb}{\left[ \begin{array}}
\newcommand{\matrixe}{\end{array} \right]}

\renewcommand{\paragraph}[1]{\noindent\textbf{#1}\quad}

%% file: sec0_abstract.tex
\begin{abstract}
\noindent
How much can we infer about a person's looks from the way they speak? In this paper, we study the task of reconstructing a facial image of a person from a short audio recording of that person speaking. We design and train a deep neural network to perform this task using millions of natural Internet/YouTube videos of people speaking.
During training, our model learns voice-face correlations that allow it to produce images that capture various physical attributes of the speakers such as age, gender and ethnicity. This is done in a self-supervised manner, by utilizing the natural co-occurrence of faces and speech in Internet videos, without the need to model attributes explicitly. 
We evaluate and numerically quantify how---and in what manner---our Speech2Face reconstructions, obtained directly from audio, resemble the true face images of the speakers.


\vspace{-3mm}
\end{abstract}


%% file: sec1_introduction.tex
\section{Introduction}
\noindent
When we listen to a person speaking without seeing his/her face, on the phone, or on the radio, we often build a mental model for the way the person looks~\cite{kamachi2003putting,smith2016matching}. There is a strong connection between speech and appearance, part of which is a direct result of the mechanics of speech production: age, gender (which affects the pitch of our voice), the shape of the mouth, facial bone structure, thin or full lips---all can affect the sound we generate. 
In addition, other voice-appearance correlations stem from the \emph{way} in which we talk: language, accent, speed, pronunciations---such properties of speech are often shared among nationalities and cultures, which can in turn translate to common physical features~\cite{denes1993speech}. 

\begin{figure}[t!]
    \centering
    \includegraphics[width=\columnwidth]{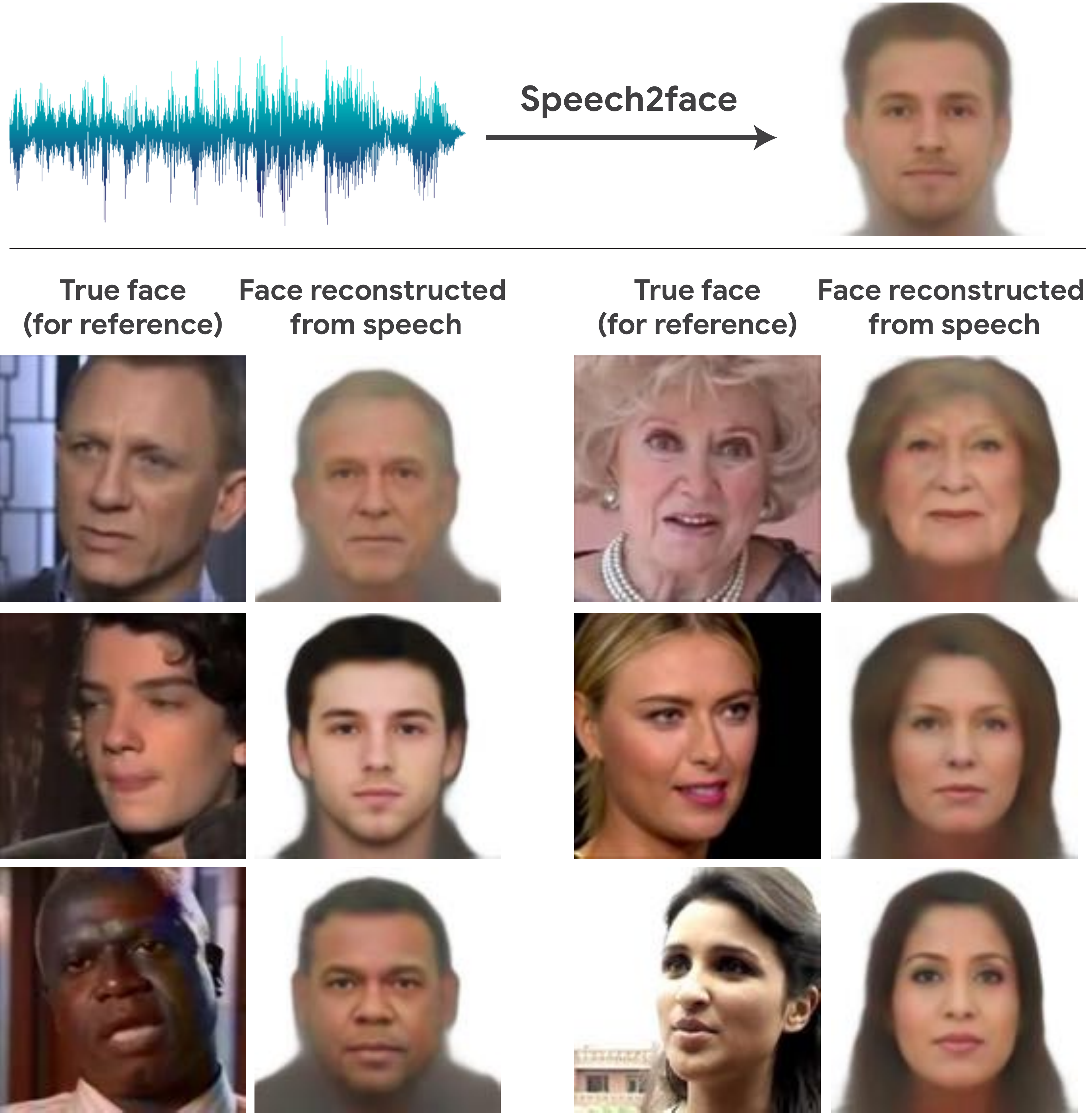}
    \caption{\emph{Top:} We consider the task of reconstructing an image of a person's face from a short audio segment of speech. \emph{Bottom:} Several results produced by our Speech2Face model, which takes only an audio waveform as input; the true faces are shown just for reference. Note that our goal is \emph{not} to reconstruct an accurate image of the person, but rather to recover characteristic physical features that are correlated with the input speech. All our results  including the input audio, are available in the supplementary material (SM).}
    \label{fig:teaser}\afterfig
\end{figure}

Our goal in this work is to study to what extent we can infer how a person looks from the way they talk. Specifically, from a short input audio segment of a person speaking, our method directly reconstructs an image of the person's face in a canonical form (i.e., frontal-facing, neutral expression). Fig.~\ref{fig:teaser} shows sample results of our method.  Obviously, there is no one-to-one matching between faces and voices. Thus, our goal is \emph{not} to predict a recognizable image of the exact face, but rather to capture dominant facial traits of the person that are correlated with the input speech.



\begin{figure*}[t!]
\vspace{-.2in}
\centering
\includegraphics[width=1\textwidth]{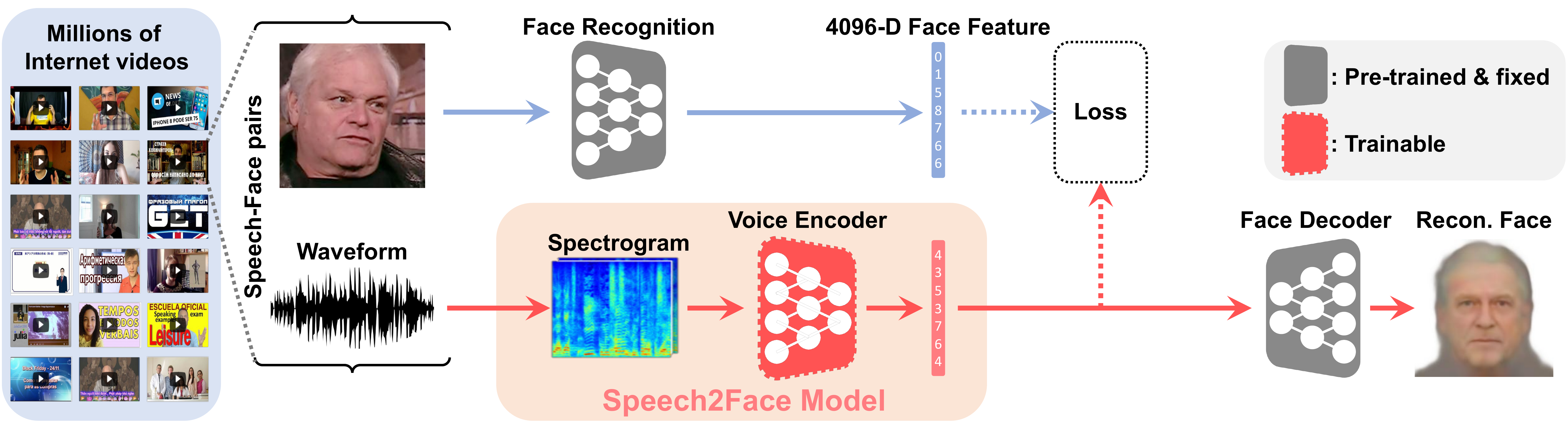}
\caption{{\bf Speech2Face model and training pipeline.} The input to our network is a complex spectrogram computed from the short audio segment of a person speaking. The output is a 4096-D face feature that is then decoded into a canonical image of the face using a pre-trained face decoder network~\protect\cite{cole2017synthesizing}. The module we train is marked by the orange-tinted box. We train the network to regress to the true face feature computed by feeding an image of the person (representative frame from the video) into a face recognition network~\cite{Parkhi15} and extracting the feature from its penultimate layer. Our model is trained on millions of speech--face embedding pairs from the AVSpeech dataset~\cite{ephrat2018looking}.}
\label{fig:pipleine}\afterfig
\end{figure*}

We design a neural network model that takes the complex
spectrogram of a short speech segment as input and predicts
a feature vector representing the face. More specifically, face information is represented by a 4096-D feature that is extracted from the penultimate layer (i.e., one layer prior to the classification layer) of a pre-trained face recognition network~\cite{Parkhi15}. We decode the predicted face feature into a canonical image of the person's face using a separately-trained reconstruction model~\cite{cole2017synthesizing}. To train our model, we use the AVSpeech dataset~\cite{ephrat2018looking}, comprised of millions of video segments from YouTube with more than 100,000 different people speaking. Our method is trained in a self-supervised manner, i.e., it simply uses the natural co-occurrence of speech and faces in videos, not requiring additional information, e.g., human annotations. 

We are certainly not the first to attempt to infer information about people from their voices. For example, predicting age and gender from speech has been widely explored~\cite{zazo2018age,hansen2015speaker,feld2010automatic,bahari2011speaker,wen2018disjoint}.
Indeed, one can consider an alternative approach to attaching a face image to an input voice by first predicting some attributes from the person's voice (e.g., their age, gender, etc.~\cite{zazo2018age}), and then either fetching an image from a database that best fits the predicted set of attributes, or using the attributes to generate an image~\cite{yan2016attribute2image}. However, this approach has several limitations. First, predicting attributes from an input signal relies on the existence of robust and accurate  classifiers and often requires ground truth labels for supervision. For example, predicting age, gender or ethnicity from speech requires building classifiers specifically trained to capture those properties. More importantly, this approach limits the predicted face to resemble only 
a predefined
set of attributes.


We aim at studying a more general, open question: what kind of facial information can be extracted from speech? Our approach of predicting full visual appearance (e.g., a face image) directly from speech allows us to explore this question without being restricted to predefined facial traits. Specifically, we show that our reconstructed face images can be used as a proxy to convey the visual properties of the person including age, gender and ethnicity.  Beyond these dominant features, our reconstructions reveal non-negligible correlations between craniofacial features~\cite{merler2019diversity} (e.g., nose structure) and voice.  This is achieved with no prior information or the existence of accurate classifiers for these types of fine geometric features.  In addition, we believe that predicting face images directly from voice may support useful applications, such as attaching a representative face to phone/video calls based on the speaker's voice.

To our knowledge, our work is the first to explore a generic (speaker independent) model for reconstructing face images directly from speech. We test our model on various speakers and numerically evaluate different aspects of our reconstructions including: how well a true face image can be retrieved based solely on an audio query; and how well our reconstructed face images agree with the true face images (unknown to the method) in terms of age, gender, ethnicity, and various craniofacial measures and ratios.

%% file: appendix.tex
\section{Ethical Considerations}
\label{sec:ethical}
\noindent
Although this is a purely academic investigation, we feel that it is important to explicitly discuss in the paper a set of ethical considerations due to the potential sensitivity of facial information.

\beforepar\paragraph{Privacy.} As mentioned, our method cannot recover the true identity of a person from their voice (i.e., an exact image of their face). 
This is because our model is trained to capture visual features (related to age, gender, etc.) that are common to \emph{many} individuals, and only in cases where there is strong enough evidence to connect those visual features with vocal/speech attributes in the data (see ``voice-face correlations'' below). As such, the model will only produce average-looking faces, with characteristic visual features that are correlated with the input speech. It will not produce images of specific individuals.

\beforepar\paragraph{Voice-face correlations and dataset bias.} Our model is designed to reveal statistical correlations that exist between facial features and voices of speakers in the training data. The training data we use is a collection of educational videos from YouTube \cite{ephrat2018looking}, and does not represent equally the entire world population. Therefore, the model---as is the case with any machine learning model---is affected by this uneven distribution of data. 

More specifically, if a set of speakers might have vocal-visual traits that are relatively uncommon in the data, then the quality of our reconstructions for such cases may degrade.  For example, if a certain language does not appear in the training data, our reconstructions will not capture well the facial attributes that may be correlated with that language.

Note that some of the features in our predicted faces may not even be physically connected to speech, for example hair color or style. However, if many speakers in the training set who speak in a similar way (e.g., in the same language) also share some common visual traits (e.g., a common hair color or style), then those visual traits may show up in the predictions. 
 
For the above reasons, we recommend that any further investigation or practical use of this technology will be carefully tested to ensure that the training data is representative of the intended user population. If that is not the case, more representative data should be broadly collected.



\beforepar\paragraph{Categories.}  In our experimental section, we mention inferred demographic categories such as ``White'' and ``Asian''. These are categories defined and used by a commercial face attribute classifier~\cite{face++}, and were only used for evaluation in this paper. Our model is not supplied with and does not make use of this information at any stage.

%% file: sec2_background.tex
\section{Related Work}

\begin{table*}[t!]
\vspace{-.1in}
\centering
\small
\resizebox{1\linewidth}{!}{
\begin{tabular}{@{}c|cgcgcgcgcgcgcgcgc@{}}
\toprule
\multirow{3}{*}{Layer}			&		&CONV			&CONV	&CONV	&			&CONV	&			&CONV	&			&CONV	&			&CONV	&CONV	&		&AVGPOOL	&		&	\\
		&Input	&RELU			&RELU	&RELU	&MAXPOOL	&RELU	&MAXPOOL	&RELU	&MAXPOOL	&RELU	&MAXPOOL	&RELU	&RELU	&CONV	&RELU		&FC		&FC	\\
			&		&BN				&BN		&BN		&			&BN		&			&BN		&			&BN		&			&BN		&BN		&		&BN			&RELU	&	\\\midrule
Channels	&2		&64				&64		&128	&--			&128	&--			&128	&--			&256	&--			&512	&512	&512	&--			&4096	&4096\\
Stride		&--		&1				&1		&1		&$2\times1$	&1		&$2\times1$	&1		&$2\times1$	&1		&$2\times1$	&1		&2		&2		&1			&1		&1	\\
Kernel size	&--		&$4\times 4$	&$4\times 4$	&$4\times 4$	&$2\times 1$	&$4\times 4$	&$2\times 1$	&$4\times 4$	&$2\times 1$	&$4\times 4$	&$2\times 1$	&$4\times 4$	&$4\times 4$	&$4\times 4$	&$\mathsf{\infty}\times 1$	&$1 \times 1$	&$1 \times 1$\\
\bottomrule
\end{tabular}
}
\vspace{1mm}
\caption{{\bf Voice encoder architecture.} The input spectrogram dimensions are 598\,$\times$\,257 (time\,$\times$\,frequency) for a 6-second audio segment (which can be arbitrarily long), with the two input channels in the table corresponding to the spectrogram's real and imaginary components.}
\label{tab:architecture}\afterfig
\end{table*}
\paragraph{Audio-visual cross-modal learning.}
The natural co-occurrence of audio and visual signals often provides rich supervision signal, without explicit labeling, also known as self-supervision~\cite{sa1994minimizing} or natural supervision~\cite{isola2015discovery}. Arandjelovi\'c and Zisserman~\cite{arandjelovic2017look} leveraged this to learn a generic audio-visual representations  by training a deep network to classify if a given video frame and a short audio clip correspond to each other.
Aytar \etal~\cite{aytar2016soundnet} proposed a student-teacher training procedure in which a well established visual recognition model was used to transfer the knowledge obtained in the visual modality to the sound modality, using unlabeled videos. Similarly, Castrejon \etal~\cite{castrejon2016learning} designed a shared audio-visual representation that is agnostic of the modality. Such learned audio-visual representations have been used for cross-modal retrieval~\cite{owens2016visually,owens2018learning,Soler2016}, sound source localization~\cite{senocak2018learning,Arandjelovic2018,owens2018audio},
and sound source separation~\cite{zhao2018the,ephrat2018looking}.
Our work utilizes the natural co-occurrence of faces and voices in Interent videos. We use a pre-trained face recognition network to transfer facial information to the voice modality.

\beforepar\paragraph{Speech-face association learning.}
The associations between faces and voices have been studied extensively in many scientific disciplines. In the domain of computer vision, different cross-modal matching methods have been proposed: a binary or multi-way classification task~\cite{nagrani2018seeing,nagrani2018learnable,shon2018noise}; metric learning~\cite{kim2018on,horiguchi2018face}; and the multi-task classification loss~\cite{wen2018disjoint}.
Cross-modal signals extracted from faces and voices have been used to disambiguate voiced and unvoiced consonants~\cite{ngiam2011multimodal,Chung2017}; to identify active speakers of a video from non-speakers therein~\cite{hoover2017putting,gebru2015tracking}; to separate mixed speech signals of multiple speakers~\cite{ephrat2018looking}; to predict lip motions from speech~\cite{ngiam2011multimodal,andrew2013deep}; or to learn the correlation between speech and emotion~\cite{albanie2018emotion}.
%
Our goal is to learn the correlations between facial traits and speech, by directly reconstructing a face image from a short audio segment.  

\beforepar \paragraph{Visual reconstruction from audio.}
Various methods have been recently proposed to reconstruct visual information from different types of audio signals. 
In a more graphics-oriented application, automatic generation of facial or body animations from music or speech has been gaining interest~\cite{Taylor2017,karras2017audio,suwajanakorn2017synthesizing,shlizerman2018audio}.
However, such methods typically parametrize the reconstructed subject a priori, and its texture is manually created or mined from a collection of textures. 
 In the context of pixel-level generative methods, Sadoughi and Busso~\cite{sadoughi2018speech} reconstruct lip motions from speech, and Wiles \etal~\cite{wiles2018x2face}  control the pose and expression of a given face using audio (or another face). While not directly related to audio, 
Yan et al.~\cite{yan2016attribute2image} and Liu and Tuzel~\cite{liu2016coupled} synthesize a face image from given facial attributes as input. Our model reconstructs a face image  directly from speech, with no additional information. Finally, Duarte~\etal\cite{wav2pix2019icassp} synthesize face images from speech using a GAN model, but their goal is to recover the true face of the speaker including  expression and pose. In contrast, our goal is to recover general facial traits, i.e., average looking faces in canonical pose and expression but capturing dominant visual attributes across many speakers.

%% file: sec3_implementation.tex
\begin{figure*}
    \centering
    \vspace{-.3cm}
    \includegraphics[width=.9\textwidth]{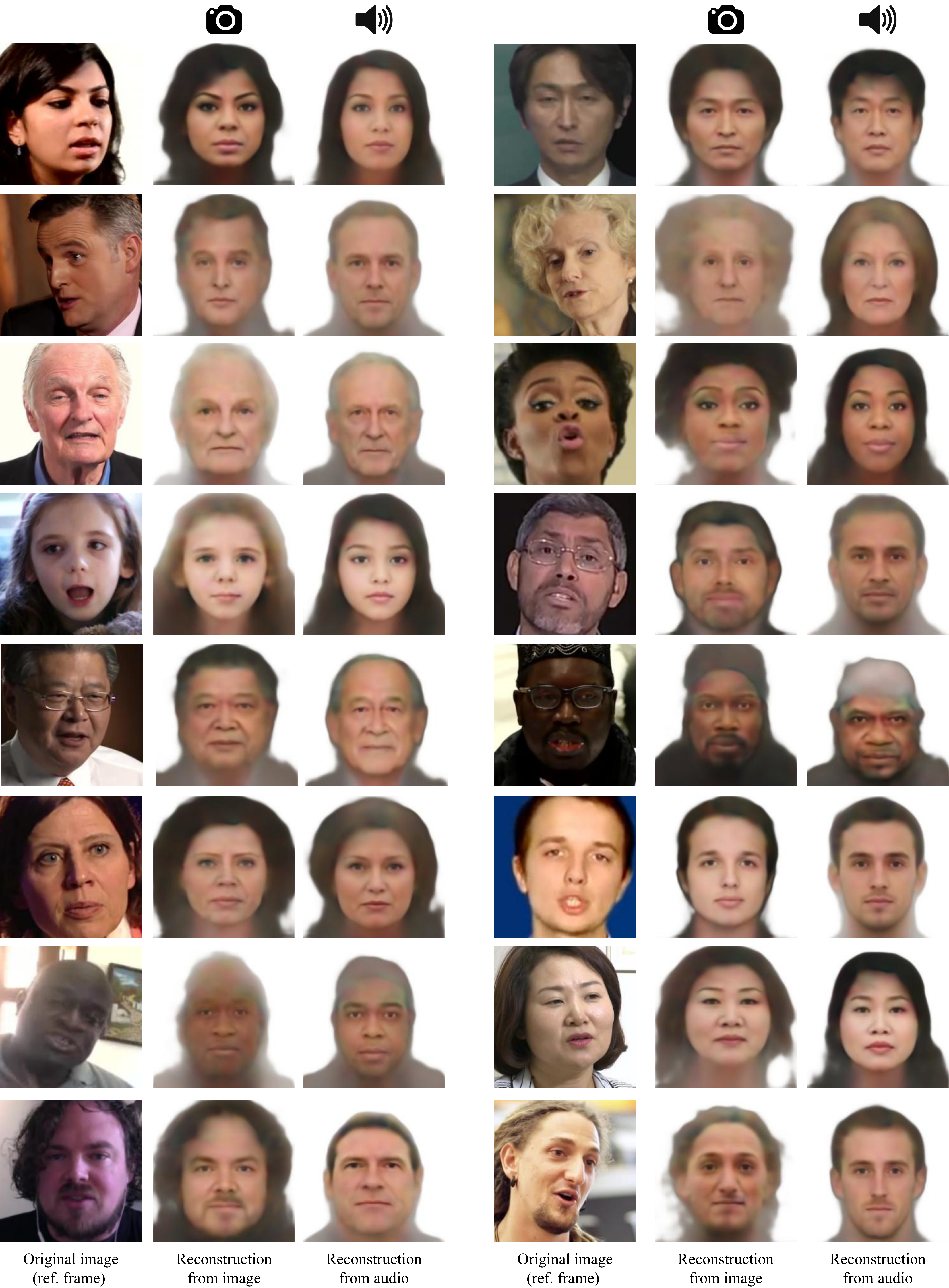}
    \caption{{\bf Qualitative results on the AVSpeech test set.} For every example (triplet of images) we show: (left) the original image, i.e., a representative frame from the video cropped around the speaker's face; (middle) the frontalized, lighting-normalized face decoder reconstruction from the VGG-Face feature extracted from the original image; (right) our Speech2Face reconstruction, computed by decoding the predicted VGG-Face feature from the audio. In this figure, we highlight successful results of our method. Some failure cases are shown in~\figref{mismatch}, and more results (including the input audio for all the examples) can be found in the SM.}
    \label{fig:av_speech_res}\afterfig
\end{figure*}

\section{Speech2Face (S2F) Model}
\label{sec:model}

\noindent
The large variability in facial expressions, head poses, occlusions, and lighting conditions in natural face images makes the design and training of a Speech2Face model non-trivial. For example, a straightforward approach of regressing from input speech to image pixels does not work; such a model has to learn to factor out many irrelevant variations in the data and to implicitly extract a meaningful internal representation of faces---a challenging task by itself.


To sidestep these challenges, we train our model to regress to a low-dimensional intermediate representation of the face.
More specifically, we utilize the VGG-Face model, a face recognition model pre-trained on a large-scale face dataset~\cite{Parkhi15}, and extract a 4096-D \emph{face feature} from the penultimate layer (\texttt{fc7}) of the network. These face features were shown to contain enough information to reconstruct the corresponding face images while being robust to many of the aforementioned variations~\cite{cole2017synthesizing}. 

Our Speech2Face pipeline, illustrated in \figref{pipleine}, consists of two main components: 1) a voice encoder, which takes a complex spectrogram of speech as input, and predicts a low-dimensional face feature that would correspond to the associated face; and 2) a face decoder, which takes as input the face feature and produces an image of the face in a canonical form (frontal-facing and with neutral expression). During training, the face decoder is fixed, and we train only the voice encoder that predicts the face feature. The voice encoder is a model we designed and trained, while 
we used a face decoder model proposed by Cole~\etal\cite{cole2017synthesizing}. We now describe both models in detail.

\begin{figure*}[h!t]
    \centering
    \includegraphics[width=0.99\textwidth]{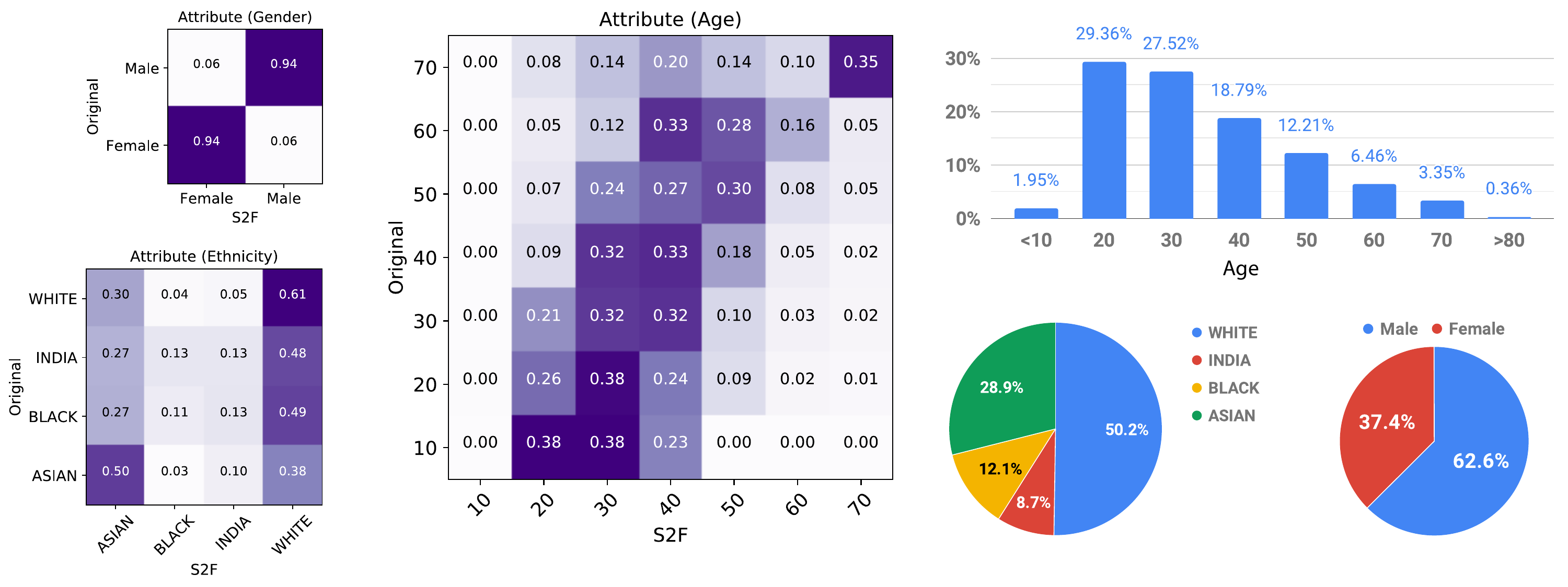}\vspace{-2mm}\\
    \hspace{1cm}{\footnotesize (a) Confusion matrices for the attributes \hspace{4.5cm} (b) AVSpeech dataset statistics}\vspace{1mm}
    \caption{{\bf Facial attribute evaluation.} (a) confusion matrices (with row-wise normalization) comparing the classification results on our Speech2Face image reconstructions (S2F) and those obtained from the original images for gender, age, and ethnicity; the stronger diagonal tendency the better performance. Ethnicity performance in (a) appears to be biased due to uneven distribution of the training set shown in~(b).
    }
    \label{fig:attr_eval}\afterfig
\end{figure*}

\beforepar\paragraph{Voice encoder network.}
Our voice encoder module is a convolutional neural network that turns the spectrogram of a short input speech into a pseudo face feature, which is subsequently fed into the face decoder to reconstruct the face image~(\figref{pipleine}).
The architecture of the voice encoder is summarized in \tabref{architecture}.
%
%
The blocks of a convolution layer, ReLU, and batch normalization~\cite{ioffe2015batch} alternate with max-pooling layers, which pool along only the temporal dimension of the spectrograms, while leaving the frequency information carried over. This is intended to preserve more of the vocal characteristics, since they are better contained in the frequency content, whereas linguistic information usually spans longer time duration~\cite{hsu2017unsupervised}.
At the end of these blocks, we apply average pooling along the temporal dimension. This allows us to efficiently aggregate information over time and makes the model applicable to input speech of varying duration.
The pooled features are then fed into two fully-connected layers to produce a 4096-D face feature.

\beforepar\paragraph{Face decoder network.}
The goal of the face decoder is to reconstruct the image of a face from a low-dimensional face feature. We opt to factor out any irrelevant variations (pose, lighting, etc.), while preserving the facial attributes. To do so, we use the face decoder model of Cole \etal~\cite{cole2017synthesizing} to reconstruct a canonical face image.
We train this model using the same face features extracted from the VGG-Face model as input to the face decoder. This model is trained separately and kept fixed during the voice encoder training.


\newcommand{\ff}{\mathbf{v}_{\!f}}
\newcommand{\vf}{\mathbf{v}_{\!s}}
\newcommand{\Lone}[2]{\left\| #1 - #2 \right\|_1}
\newcommand{\Ltwo}[2]{\left\| #1 - #2 \right\|_2^2}
\newcommand{\lastenc}{f_\texttt{VGG}}
\newcommand{\firstdec}{f_\texttt{dec}}
\newcommand{\Ldistill}{L_\mathrm{distill}}

\beforepar\paragraph{Training.}
Our voice encoder is trained in a self-supervised manner, using the natural co-occurrence of a speaker's speech and facial images in videos.
To this end, we use the AVSpeech dataset~\cite{ephrat2018looking}, a large-scale ``in-the-wild'' audiovisual dataset of people speaking.
A single frame containing the speaker's face is extracted from each video clip and fed to the VGG-Face model to extract the 4096-D feature vector, $\ff$. This serves as the supervision signal for our voice encoder---the feature, $\vf$, of our voice encoder is trained to predict $\ff$.
%

A natural choice for the loss function would be the $L_1$ distance between the features: $\Lone{\ff}{ \vf}$. However, we found that the training undergoes slow and unstable progression 
with this loss alone. 
To stabilize the training, we introduce additional loss terms, motivated by Castrejon et al.~\cite{castrejon2016learning}. Specifically, we additionally penalize the difference in the activation of the \emph{last} layer of the face encoder, $\lastenc : \mathbb{R}^{4096} \rightarrow \mathbb{R}^{2622}$, i.e., \texttt{fc8} of VGG-Face, and that of the \emph{first} layer of the face decoder, $\firstdec : \mathbb{R}^{4096}{\rightarrow} \mathbb{R}^{1000}$, which are pre-trained and fixed during training the voice encoder. We feed both our predictions and the ground truth face features to these layers to calculate the losses. The final loss is:\vspace{-2mm}
\begin{eqnarray}
    \mathcal{L}_\mathrm{total} =&\hspace{-5mm}& \, \Lone{\firstdec(\ff)}{\firstdec(\vf)} + \lambda_{1} 
    \Ltwo{\tfrac{\ff}{\|\ff\|}}{\tfrac{\vf}{\|\vf\|}} \nonumber\\
    &\hspace{-5mm}& \, + \lambda_{2} \, \Ldistill\left(\lastenc(\ff), \lastenc(\vf)\right),
\label{eq:total_loss}
\end{eqnarray}
where $\lambda_{1}{=}0.025$ and $\lambda_{2}{=}200$.
$\lambda_1$ and $\lambda_2$ are tuned such that the gradient magnitude of each term with respect to $\vf$ are within a similar scale at an early iteration (we measured at the 1000th iteration).
The knowledge distillation loss $\Ldistill(\mathbf{a},\mathbf{b}) = -\!\sum_i p_{(i)}(\mathbf{a})\log p_{(i)}(\mathbf{b})$, where $p_{(i)}(\mathbf{a}) = \tfrac{\exp(a_i/T)}{\sum_j{\exp(a_j/T)}}$, is used as an alternative of the cross entropy loss, which encourages the output of a network to approximate the output of another~\cite{hinton2015distilling}. $T{=}2$ is used as recommended by the authors, which makes the activation smoother.
We found that enforcing similarity over these additional layers stabilized and sped up the training process, in addition to a slight improvement in the resulting quality.

\beforepar\paragraph{Implementation details.}
We use up to 6 seconds of audio taken from the beginning of each video clip in AVSpeech. If the video clip is shorter than 6 seconds, we repeat the audio such that it becomes at least 6-seconds long. The audio waveform is resampled at 16\,kHz and only a single channel is used. Spectrograms are computed similarly to Ephrat et al.~\cite{ephrat2018looking} by taking STFT with a Hann window of 25\,mm, the hop length of 10\,ms, and 512 FFT frequency bands. Each complex spectrogram $S$ subsequently goes through the power-law compression, resulting $\mathrm{sgn}(S)|S|^{0.3}$ for real and imaginary independently, where $\mathrm{sgn}(\cdot)$ denotes the signum.
%
%
We run the CNN-based face detector from \texttt{Dlib}~\cite{dlib09}, crop the face regions from the frames, and resize them to 224\,$\times$\,224 pixels. The VGG-Face features are computed from the resized face images. The computed spectrogram and VGG-Face feature of each segment are collected and used for training. The resulting training and test sets include 1.7 and 0.15 million spectra--face feature pairs, respectively.
Our network is implemented in TensorFlow and optimized by ADAM~\cite{kingma2014adam} with $\beta_1 = 0.5$, $\epsilon = 10^{-4}$, the learning rate of 0.001 with the exponentially decay rate of 0.95 at every 10,000 iterations, and the batch size of 8 for 3 epochs. 

%
%
%
%
%
%
%
%
%
%
%
%
%

%% file: sec4_results.tex
\section{Results}

\noindent
We test our model both qualitatively and quantitatively on the AVSpeech dataset~\cite{ephrat2018looking} and the VoxCeleb dataset~\cite{Nagrani2017}. Our goal is to gain insights and to quantify how---and in which manner---our Speech2Face reconstructions resemble the true face images. 

Qualitative results on the AVSpeech test set are shown in \figref{av_speech_res}. For each example, we show the true image of the speaker for reference (unknown to our model), the face reconstructed from the \emph{face} feature (computed from the true image) by the face decoder (Sec.~\ref{sec:model}), and the face reconstructed from a 6-seconds audio segment of the person's speech, which is our Speech2Face result.
While looking somewhat like average faces, our Speech2Face reconstructions 
capture rich physical information about the speaker, such as their age, gender, and ethnicity. The predicted images also capture additional properties like the shape of the face or head (e.g., elongated vs.\ round), which we often find consistent with the true appearance of the speaker; see the last two rows in~\figref{av_speech_res} for instance. 


\subsection{Facial Features Evaluation}
\noindent
We quantify how well different facial attributes are being captured in our Speech2Face reconstructions and test different aspects of our model.

\beforepar\paragraph{Demographic attributes.} We use Face++~\cite{face++}, a leading commercial service for computing facial attributes. Specifically, we evaluate and compare age, gender, and ethnicity, by running the Face++ classifiers on the original images and our Speech2Face reconstructions. The Face++ classifiers return either ``male'' or ``female'' for gender, a continuous number for age, and one of the four values, ``Asian'', ``black'', ``India'', or ``white'', for ethnicity.\footnote{We directly refer to the Face++ labels, which are not our terminology.}


\begin{figure}
\centering
\includegraphics[width=0.99\linewidth]{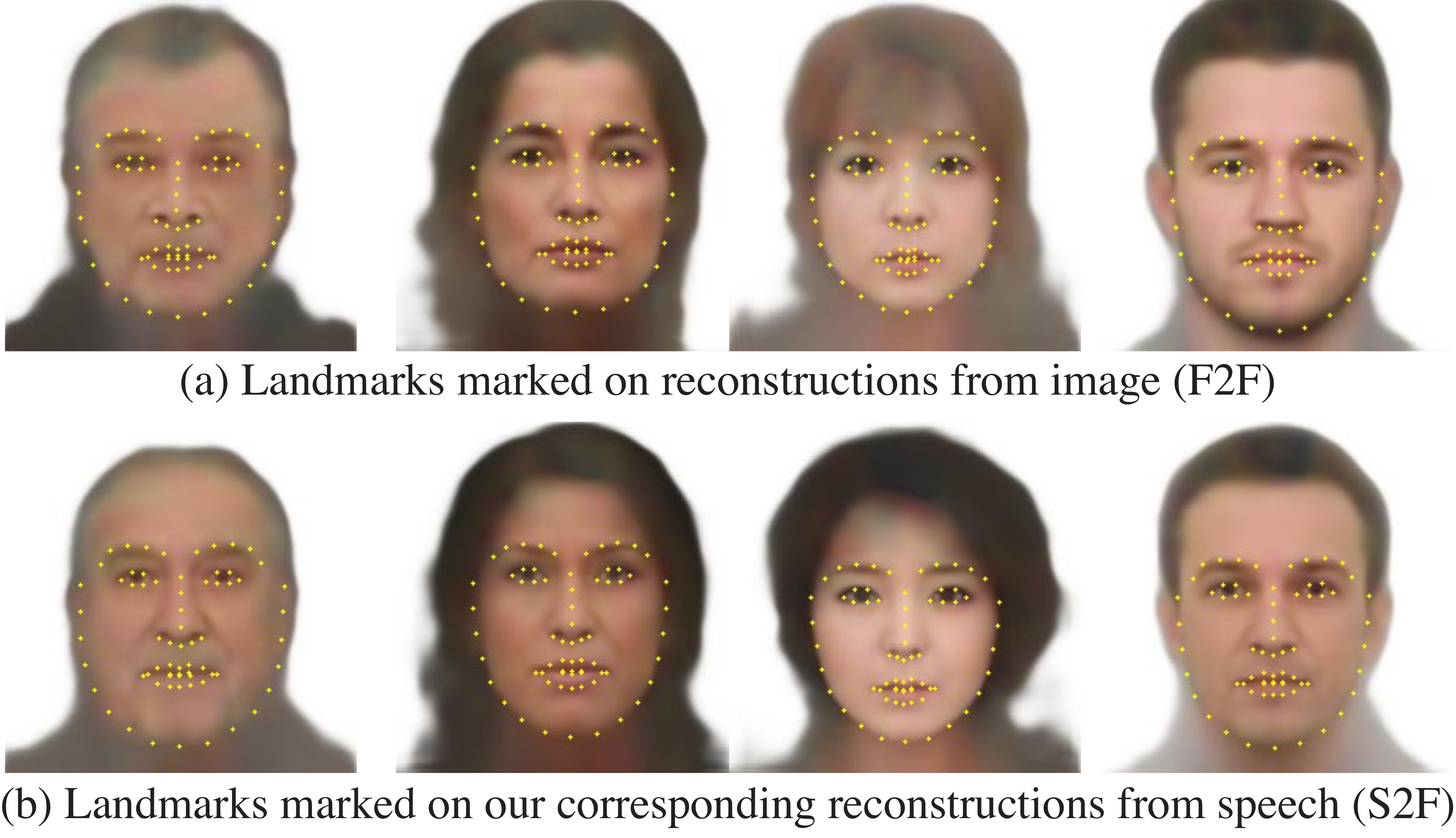}
\vspace{.1cm}
\resizebox{0.85\linewidth}{!}{
\begin{tabular}{lll}
\toprule
Face measurement \hspace{-0.2cm}  & \hspace{-0.2cm} Correlation & $p$-value\\
\midrule
Upper lip height & 0.16  & $p < 0.001$ \\
Lateral upper lip heights & 0.26 & $p < 0.001$ \\
Jaw width & 0.11 & $p < 0.001$ \\
Nose height & 0.14 & $p < 0.001$  \\
Nose width & 0.35 & $p < 0.001$ \\
Labio oral region & 0.17 & $p < 0.001$ \\
Mandibular idx & 0.20 & $p < 0.001$ \\
Intercanthal idx & 0.21 & $p < 0.001$ \\
Nasal index & 0.38 & $p < 0.001$\\
Vermilion height idx & 0.29 & $p < 0.001$ \\
Mouth face with idx &  0.20 & $p < 0.001$\\
Nose area & 0.28 & $p < 0.001$ \\\midrule
Random baseline & 0.02 & --
\\
\bottomrule 
\end{tabular}
}\\
{\scriptsize (c) Pearson correlation coefficient}
\vspace{.1cm}
\caption{\textbf{Craniofacial features.} We measure the  correlation between craniofacial features extracted from (a) face decoder reconstructions from the original image (F2F), and (b) features extracted from our corresponding Speech2Face reconstructions (S2F); the features are computed from detected facial landmarks, as described in~\cite{merler2019diversity}. The table reports Pearson correlation coefficient and statistical significance computed over 1,000 test images for each feature. Random baseline is computed for ``Nasal index'' by comparing random pairs of F2F reconstruction (a) and S2F reconstruction (b).}
\label{fig:face_eval}
\afterfig
\end{figure}

Fig.~\ref{fig:attr_eval}(a) shows confusion matrices for each of the attributes, comparing the attributes inferred from the original images with those inferred from our Speech2Face reconstructions (S2F). See the supplementary material for similar evaluations of our face-decoder reconstructions from the images (F2F). As can be seen, for age and gender the classification results are highly correlated. For gender, there is an agreement of 94$\%$ in male/female labels between the true images and our reconstructions from speech.
For ethnicity, there is a good correlation on the ``white'' and ``Asian'', but we observe less agreement on ``India'' and ``black''. We believe this is because those classes have a smaller representation in the data (see statistics we computed on AVSpeech in Fig.~\ref{fig:attr_eval}(b)). The performance can potentially be improved by leveraging the statistics to balance the training data for the voice encoder model, which we leave for future work.

\beforepar\paragraph{Craniofacial attributes.}
We evaluated craniofacial measurements commonly used in the literature, for capturing ratios and distances in the face~\cite{merler2019diversity}. For each such measurement, we computed the correlation between F2F 
(Fig.~\ref{fig:face_eval}(a)), and our corresponding 
S2F
reconstructions (Fig.~\ref{fig:face_eval}(b)). Face landmarks were computed using the DEST library~\cite{DEST}. Note that this evaluation is made possible because we are working with normalized faces (neutral expression, frontal-facing), thus differences between the facial landmarks' positions reflect geometric craniofacial changes. Fig.~\ref{fig:face_eval}(c) shows the Pearson correlation coefficient for several measures, computed over 1,000 random samples from the AVSpeech test set.  As can be seen, there is 
statistically significant
(i.e., $p<0.001$) positive correlation for several measurements. In particular, the highest correlation is measured for the nasal index (0.38) and nose width (0.35), the features indicative of nose structures that may affect a speaker's voice.


\begin{table}
\centering
\footnotesize
\resizebox{0.95\linewidth}{!}{
\begin{tabular}{c|ccc}
\toprule
{\bf Length} & {\bf cos (deg)}   & $\mathbf{L_2}$          & $\mathbf{L_1}$          \\  \midrule
3 seconds         & $48.43 \pm 6.01$  & $0.19 \pm 0.03$    & $9.81 \pm 1.74$  \\
6 seconds         & $45.75 \pm 5.09$  & $0.18 \pm 0.02$    & $9.42 \pm 1.54$   \\
\bottomrule
\end{tabular}
}
\vspace{1mm}

\caption{{\bf Feature similarity.} We measure the similarity between our features predicted from speech and the corresponding face features computed on the true images of the speakers. We report average cosine, $L_2$ and $L_1$ distances over 5000 random samples from the AVSpeech test set, using 3- and 6-second audio segments.}
\label{tab:similarity}  \afterfig
\end{table}

\begin{figure}
    \centering
    \includegraphics[width=1\linewidth]{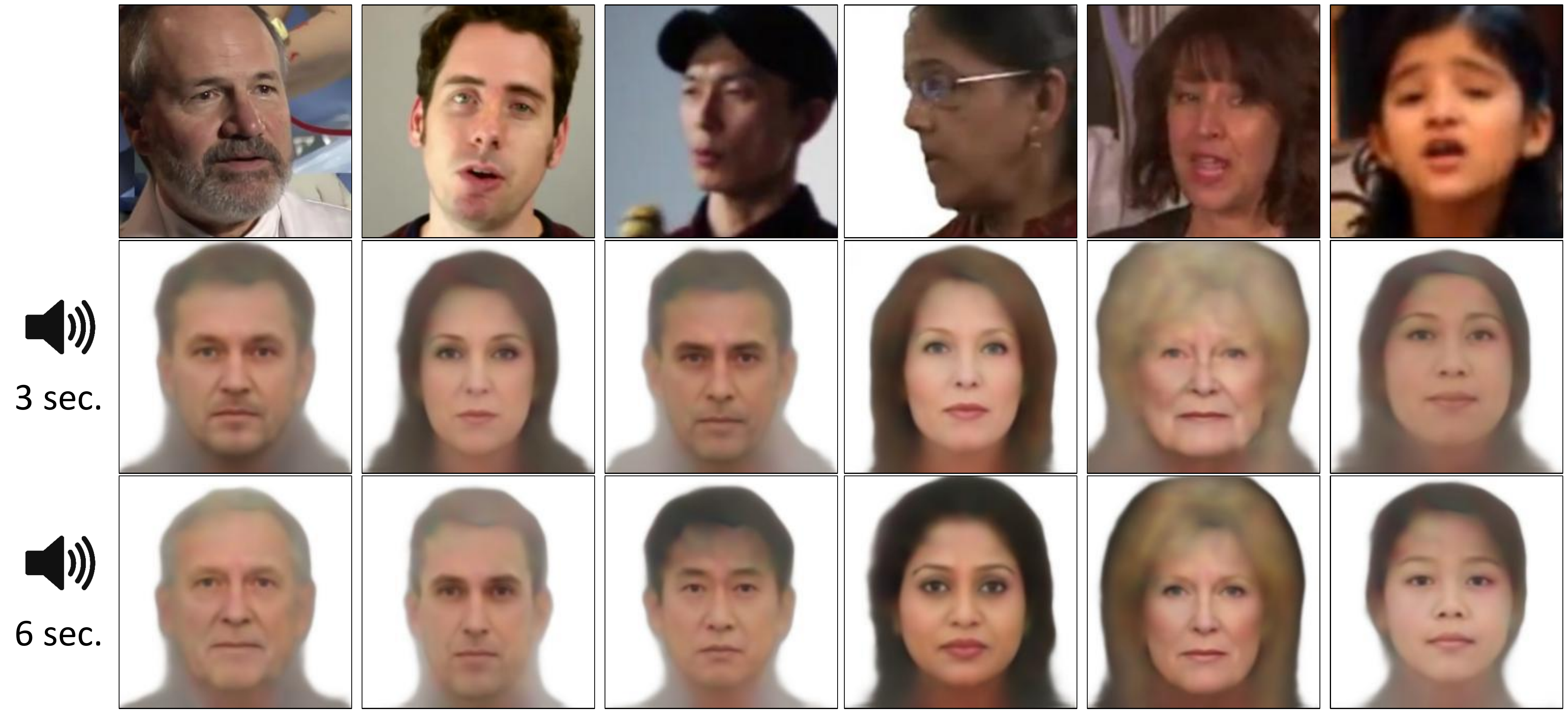}
    \caption{{\bf The effect of input audio duration.} We compare our face reconstructions when using 3-second (middle row) and 6-second (bottom row) input voice segments at test time (in both cases we use the same model, trained on 6-second segments). The top row shows representative frames from the videos for reference. With longer speech duration the reconstructed faces capture the facial attributes better.}
    \label{fig:3sec_vs_6sec_test}\afterfig
\end{figure}

\beforepar\paragraph{Feature similarity.} We test how well a person can be recognized from on the face features predicted from speech. We first directly measure the cosine distance between our predicted features and the true ones obtained from the original face image of the speaker. Table~\ref{tab:similarity} shows the average error over 5,000 test images, for the predictions using 3s and 6s audio segments. The use of longer audio clips exhibits consistent improvement in all error metrics; this further evidences the qualitative improvement we observe in Fig.~\ref{fig:3sec_vs_6sec_test}.

\begin{table}[t]
\centering
\resizebox{0.9\linewidth}{!}{
\footnotesize
\begin{tabular}{c@{\hskip 3mm}c@{\hskip 1mm}|rrrr}
\toprule
\bf Duration  &	\bf Metric	&  $\mathbf{R@1}$	& $\mathbf{R@2}$	& $\mathbf{R@5}$ & 	$\mathbf{R@10}$ \\\midrule
\bf 3 sec	&   $\mathbf{L_2}$    &   5.86	& 10.02	&   18.98	&   28.92\\
\bf 3 sec	&   $\mathbf{L_1}$    &   6.22	& 9.92	&   18.94	&   28.70\\
\bf 3 sec	&   \bf cos     &   8.54	& 13.64	&   24.80	&   38.54\\\midrule
\bf 6 sec	&   $\mathbf{L_2}$    &   8.28	& 13.66	&   24.66	&   35.84\\
\bf 6 sec	&   $\mathbf{L_1}$    &   8.34	& 13.70	&   24.66	&   36.22\\
\bf 6 sec	&   \bf cos	    &   \bf 10.92	& \bf 17.00	& \bf 30.60	&  \bf 45.82\\\midrule
\multicolumn{2}{c|}{Random} & 1.00 & 2.00 & 5.00 & 10.00 \\
\bottomrule
\end{tabular}
}
\vspace{1mm}
\caption{{\bf S2F$\rightarrow$Face retrieval performance.}
We measure retrieval performance by recall at $K$ ($R@K$, in $\%$), which indicates the chance of retrieving the true image of a speaker within the top-$K$ results. We used a database of 5,000 images for this experiment; see ~\figref{retrieval_qual} for qualitative results. The higher the better. Random chance is presented as a baseline.}
\label{tab:retrieval_quan}\afterfig
\end{table}

\begin{figure}
\centering
\resizebox{1.0\linewidth}{!}{
\begin{tabular}{@{\hskip 0mm}c@{\hskip 0mm}@{\hskip 1mm}c@{\hskip 0mm}}
\includegraphics[height=0.95\linewidth]{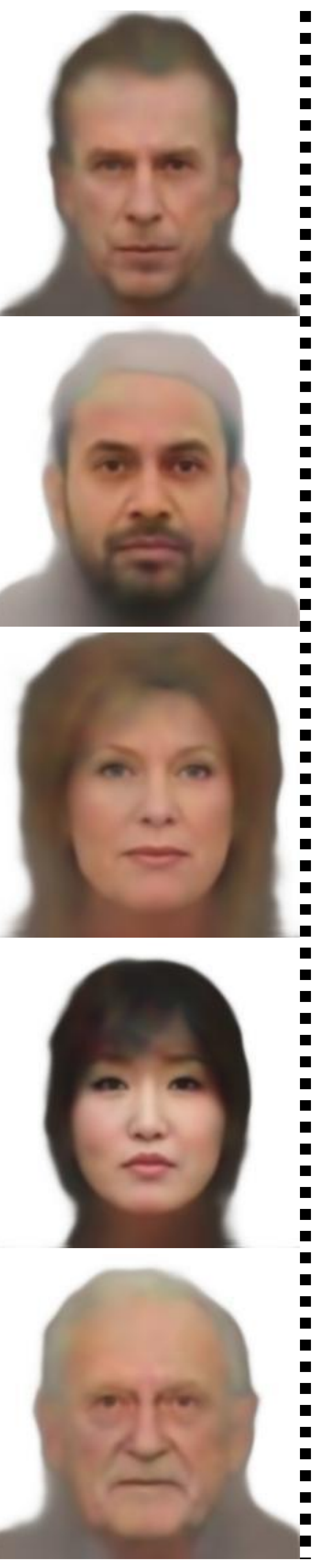}&
\includegraphics[height=0.95\linewidth]{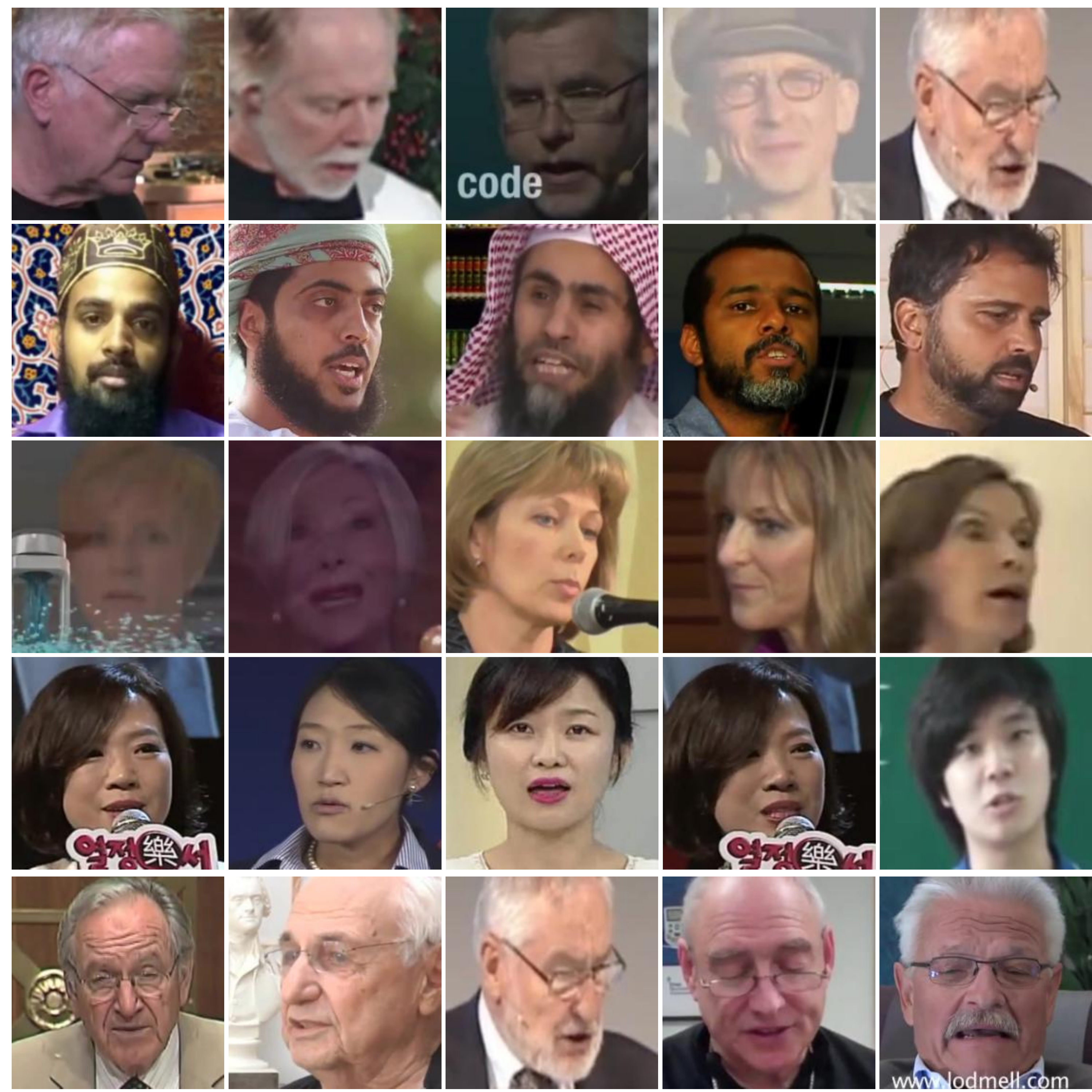}\\
S2F recon. & Retrieved top-5 results
\end{tabular}
}
\vspace{1mm}
    \caption{{\bf S2F$\rightarrow$Face retrieval examples.} 
    We query a database of 5,000 face images by comparing our Speech2Face prediction of input audio to all VGG-Face face features in the database (computed directly from the original faces). For each query, we show the top-5 retrieved samples. 
    The last row is an example where 
    the true face was not among the top results, but still shows visually close results to the query.
    More results are available in the SM.
    }
    \label{fig:retrieval_qual}
\afterfig
\end{figure}

We further evaluated how accurately we can retrieve the true speaker from a database of face images. To do so, we take the speech of a person to predict the feature using our Speech2Face model, and query it by computing its distances to the \emph{face} features of all face images in the database. We report the retrieval performance by measuring the recall at $K$, i.e., the percentage of time the true face is retrieved within the rank of $K$.
Table~\ref{tab:retrieval_quan} shows the computed recalls for varying configurations. In all cases, the cross-modal retrieval using our model shows a significant performance gain compared to the random chance.
It also shows that a longer duration of the input speech noticeably improves the performance.
In \figref{retrieval_qual}, we show several examples of 5 nearest faces such retrieved, which demonstrate the consistent facial characteristics that are being captured by our predicted face features. 

\beforepar\paragraph{t-SNE visualization for learned feature analysis.}
To gain more insights on our predicted features, we present 2-D t-SNE plots~\cite{maaten2008visualizing} of the features in the SM.

\subsection{Ablation Studies}

\ifdefined\arxivversion

\input{sec4add2_results.tex}

\fi

\begin{figure}
    \centering
    \includegraphics[width=\columnwidth]{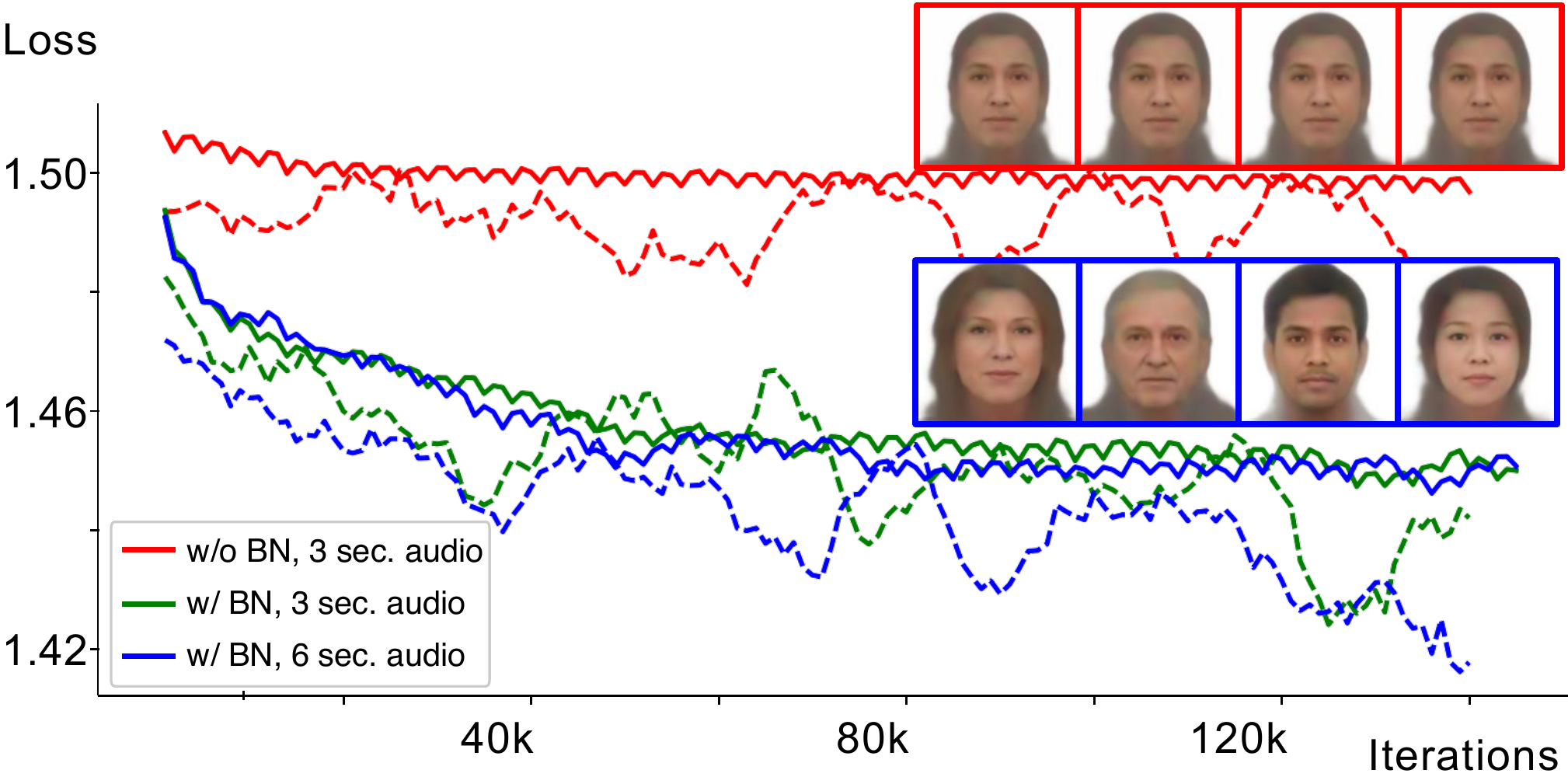}
    \caption{{\bf Training convergence patterns.}
    BN denotes batch normalization. The red and green curves are obtained by using 3- and 6-second audio clips as input during training, respectively (dashed line: training loss; solid line: validation loss). The face thumbnails show reconstructions from models trained with and without BN.}
    \label{fig:conv_graph}
    \afterfig
\end{figure}

\paragraph{The effect of audio duration and batch normalization.}
We tested the effect of the duration of the input audio during both the train and test stages. Specifically, we trained two models with 3- and 6-second speech segments. We found that during the training time, the audio duration has an only subtle effect on the convergence speed, without much effect on the overall loss and the quality of reconstructions (Fig.~\ref{fig:conv_graph}).  However, we found that feeding longer speech as input at \emph{test} time leads to improvement in reconstruction quality, that is, reconstructed faces capture the personal attributes better, regardless of which of the two models are used. Fig.~\ref{fig:3sec_vs_6sec_test} shows several qualitative comparisons, which are also consistent with the quantitative evaluations in Tables~\ref{tab:similarity} and \ref{tab:retrieval_quan}.


Fig.~\ref{fig:conv_graph} also shows the training curves w/ and w/o Batch Normalization (BN). As can be seen, without BN the reconstructed faces converge to an average face. With BN the  results contain much richer facial information.

\begin{figure}
    \centering
    \includegraphics[width=0.9\columnwidth]{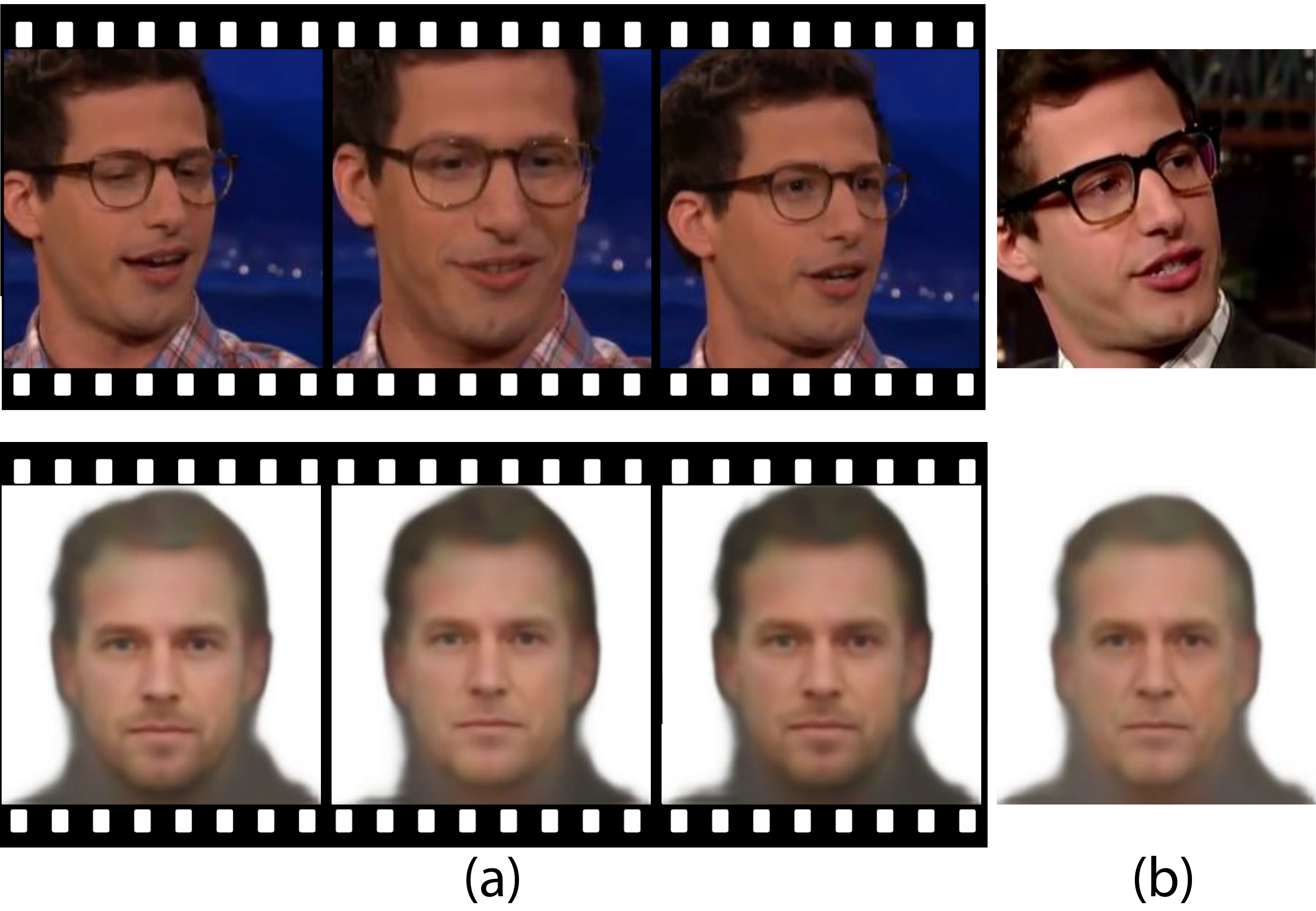}\vspace{-1mm}
    \caption{{\bf Temporal and cross-video consistency.}\,Face reconstruction from different speech segments of the same person taken from different parts within (a) the same or from (b) a different video.}
    \label{fig:temporal_consist}\afterfig
\end{figure}

\begin{figure}
\centering
\resizebox{1.00\linewidth}{!}{
\begin{tabular}{@{\hskip 0mm}p{0.48\linewidth}@{\hskip 1mm}|@{\hskip 1mm}p{0.48\linewidth}@{\hskip 0mm}}
\includegraphics[width=1\linewidth]{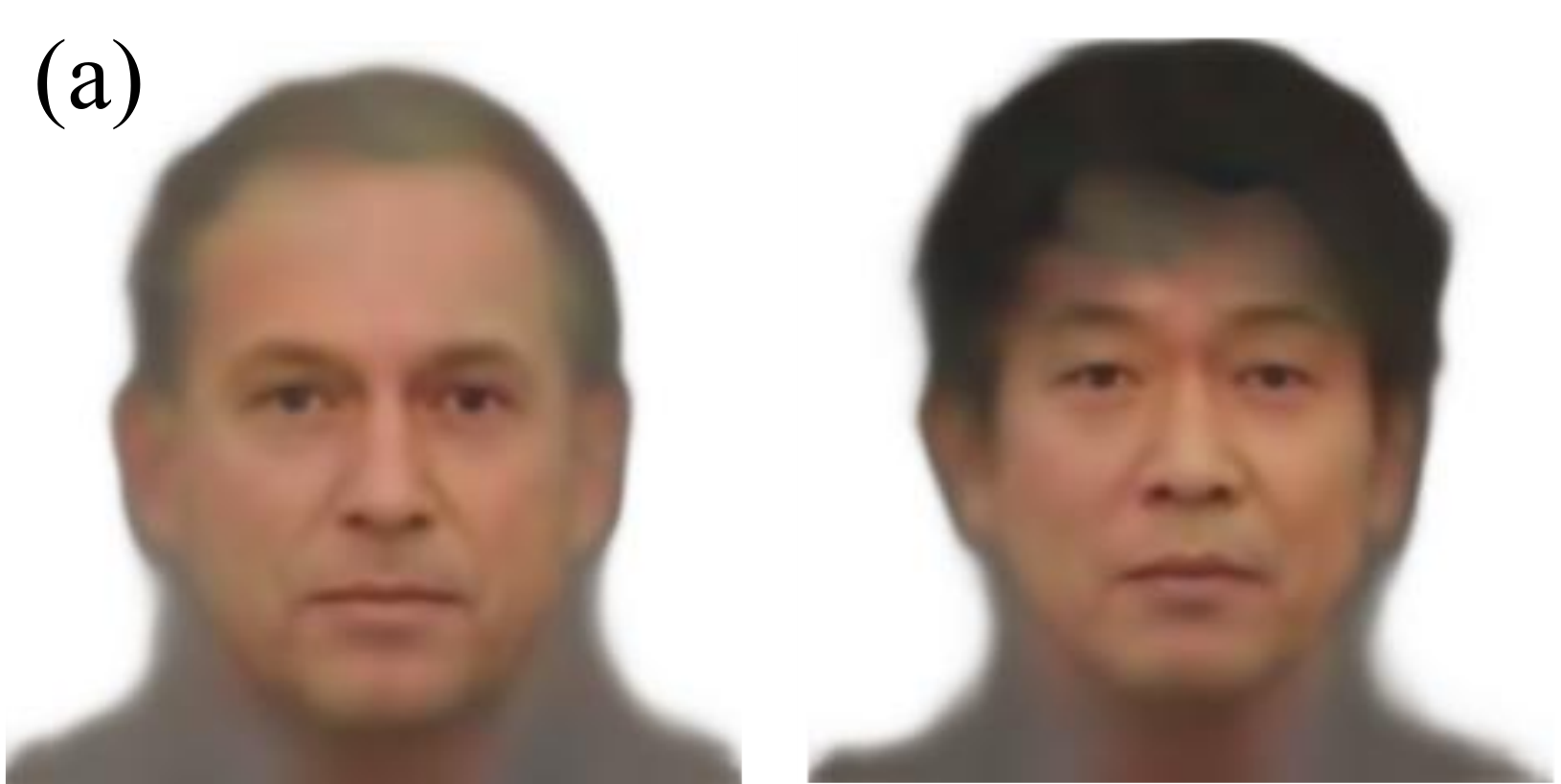}&
\includegraphics[width=1\linewidth]{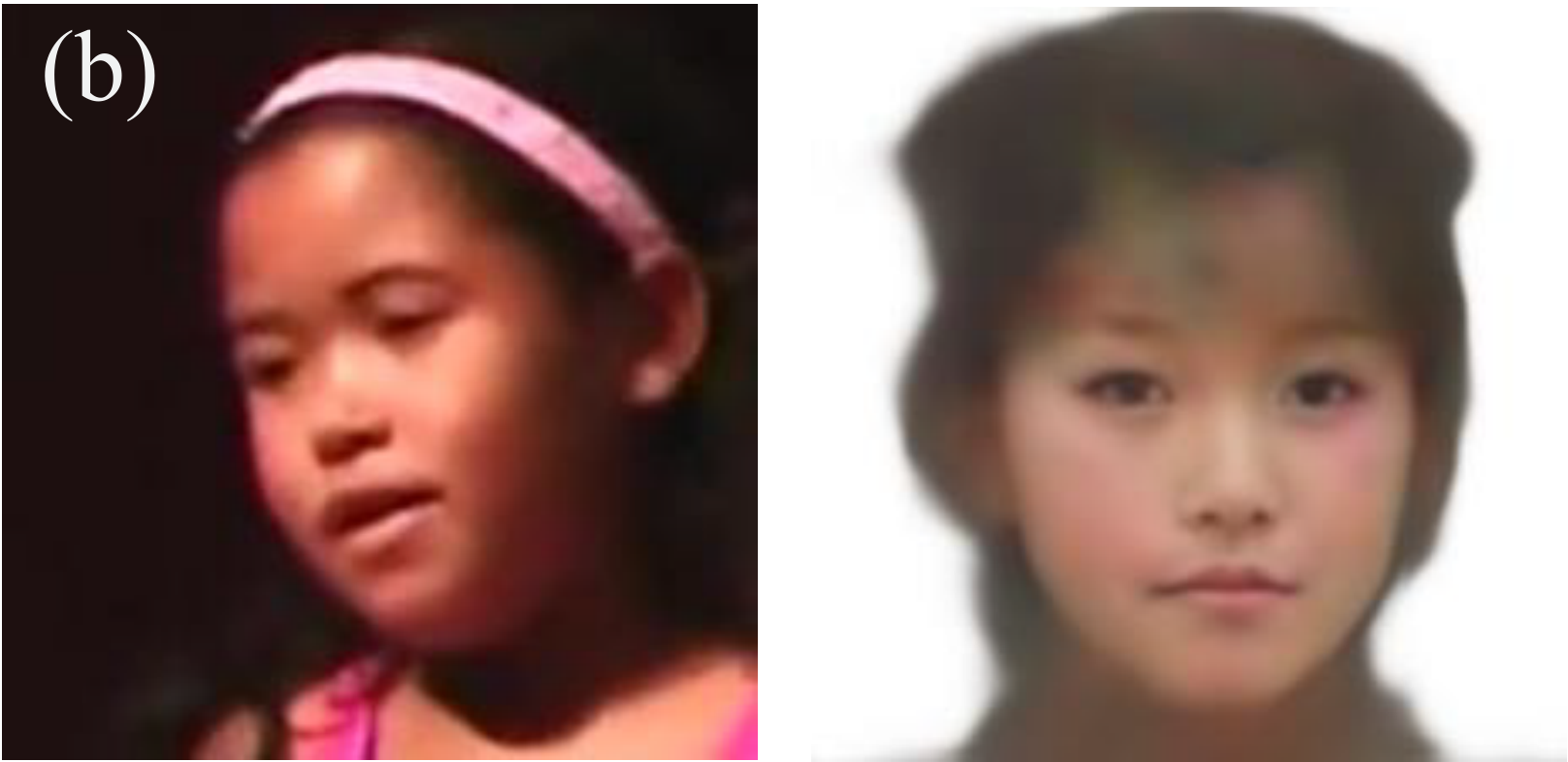}\\
\scriptsize An Asian male speaking in English (left) \& Chinese (right) &
\scriptsize An Asian girl speaking in English
\end{tabular}
}
    \caption{{\bf The effect of language.}  We notice mixed performance in terms of the ability of the model to handle languages and accents. 
    (a) A sample case of language-dependent face reconstructions.
    (b) A sample case that successfully factors out the language.
    }
    \label{fig:lang}\afterfig
\end{figure}

\begin{figure}[b]
\vspace{-5mm}
    \centering
    \includegraphics[width=0.98\columnwidth]{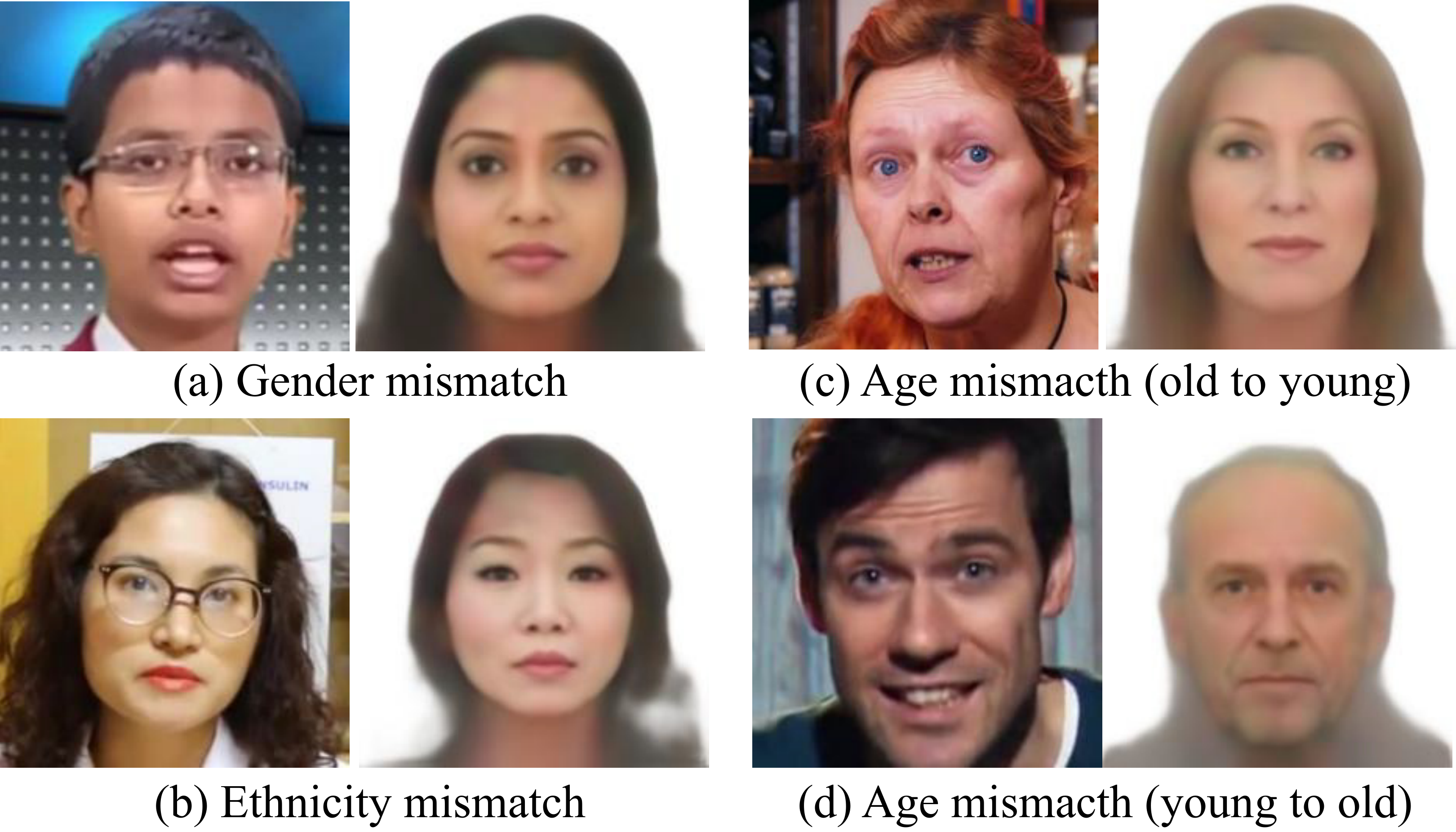}
    \caption{{\bf Example failure cases.} (a) High-pitch male voice, e.g., of kids, may lead to a face image with female features. (b) Spoken language does not match ethnicity. (c-d) Age mismatches.}
    \label{fig:mismatch}
\end{figure}

\beforepar\paragraph{Additional observations and limitations.}
In Fig.~\ref{fig:temporal_consist}, we infer faces from different speech segments of the same person, taken from different parts within the same video, and from a different video, in order to test the stability of our Speech2Face reconstruction. The reconstructed face images are consistent within and between the videos. We show more such results in the SM.

To qualitatively test the effect of language and accent, we probe the model with an Asian male example speaking the same sentence in English and Chinese (Fig.~\ref{fig:lang}(a)).
While having the same reconstructed face in both cases would be ideal, the model inferred different faces based on the spoken language. However, in other examples, e.g., Fig.~\ref{fig:lang}(b), the model was able to successfully factor out the language, reconstructing a face with Asian features even though the girl was speaking in English with no apparent accent (the audio is available in the SM). In general, we observed mixed behaviors and a more thorough examination is needed to determine to which extent the model relies on language.

More generally, the ability to capture the latent attributes from speech, such as age, gender, and ethnicity, depends on several factors such as accent, spoken language, or voice pitch. Clearly, in some cases, these vocal attributes would not match the person's appearance. Several such typical speech-face mismatch examples are shown in Fig.~\ref{fig:mismatch}.







\begin{figure}[b]
\vspace{-2mm}
    \centering
    \includegraphics[width=0.72\columnwidth]{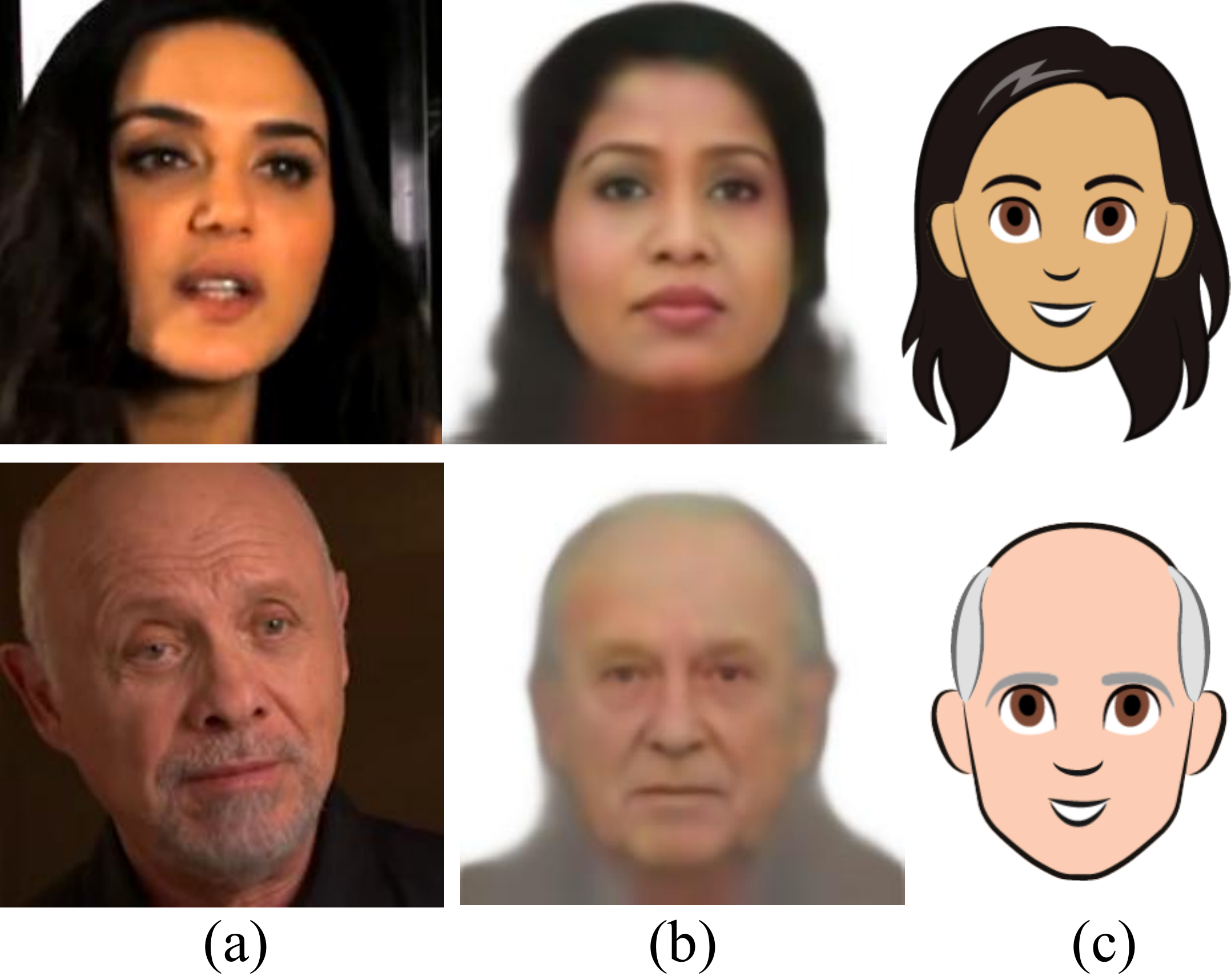}
    \caption{{\bf Speech-to-cartoon.} Our reconstructed faces from audio (b) can be re-rendered as cartoons (c) using existing tools, such as the personalized emoji app available in Gboard, the keyboard app in Android phones~\cite{face2cartoon}.
    (a) The true images of the person are shown for reference.
    }
    \label{fig:speech2cartoon}
\end{figure}

\subsection{Speech2cartoon}
\noindent
Our face images reconstructed from speech may be used for generating personalized cartoons of speakers from their voices, as shown in Fig.~\ref{fig:speech2cartoon}. We use Gboard, the keyboard app available on Android phones, which is also capable of analyzing a selfie image to produce a cartoon-like version of the face~\cite{face2cartoon}. As can be seen, our reconstructions capture the facial attributes well enough for the app to work. Such cartoon re-rendering of the face may be useful as a visual representation of a person during a phone or a video-conferencing call, when the person's identity is unknown or the person prefers not to share his/her picture. Our reconstructed faces may also be used directly, to assign faces to machine-generated voices used in home devices and virtual assistants.

%% file: sec4add2_results.tex
\begin{figure}
\newcommand{\complossimga}[1]{#10032Stevie_Wonder_Q5CakG-6vIY_0000007.jpg}
\newcommand{\complossimgb}[1]{#11013Raini_Rodriguez_qlMFZfWY-PE_0000001.jpg}
\newcommand{\complossimgc}[1]{#11115Loan_Chabanol_wR579s6AbqE_0000001.jpg}
\newcommand{\complossimgd}[1]{#10020Adam_Beach_vy8sQ82o0fM_0000003.jpg}
\newcommand{\complossimge}[1]{#10014Pablo_Schreiber_QKKD6tBn-UI_0000012.jpg}
\def\patha{figures/comploss/ori/}
\def\pathb{figures/comploss/300k_imgloss/}
\def\pathc{figures/comploss/300k_1kloss/}
\def\pathd{figures/comploss/500k_imgloss/}
\def\pathe{figures/comploss/500k_1kloss/}
\def\complosswidth{0.2\linewidth}
\newcommand{\putimg}[1]{\includegraphics[width=\complosswidth]{#1}}

{
    \resizebox{\linewidth}{!}{
    \renewcommand{\arraystretch}{0.3} 
    \scriptsize
        \begin{tabular}{@{}c|@{}c@{}c@{}|@{}c@{}c@{}}
        Iteration & \multicolumn{2}{@{}c@{}|@{}}{300k iter.} & \multicolumn{2}{c}{500k iter.}\\
        \midrule
        Original image & & & &\\
        (ref. frame) & Pixel loss & Full loss & Pixel loss & Full loss\\
        \putimg{\complossimga{\patha}} & \putimg{\complossimga{\pathb}} & \putimg{\complossimga{\pathc}} & \putimg{\complossimga{\pathd}} & \putimg{\complossimga{\pathe}}\\
        \putimg{\complossimgb{\patha}} & \putimg{\complossimgb{\pathb}} & \putimg{\complossimgb{\pathc}} & \putimg{\complossimgb{\pathd}} & \putimg{\complossimgb{\pathe}}\\
        \putimg{\complossimgc{\patha}} & \putimg{\complossimgc{\pathb}} & \putimg{\complossimgc{\pathc}} & \putimg{\complossimgc{\pathd}} & \putimg{\complossimgc{\pathe}}\\
        \putimg{\complossimgd{\patha}} & \putimg{\complossimgd{\pathb}} & \putimg{\complossimgd{\pathc}} & \putimg{\complossimgd{\pathd}} & \putimg{\complossimgd{\pathe}}\\
        \end{tabular}
    }
}
{\caption[]{
\textbf{Comparison to a pixel loss.} The results obtained with an $L_1$ loss on the output image and our full loss (Eq.~\ref{eq:total_loss}) are shown after 300k and 500k training iterations (indicating convergence).
    }\label{fig:comploss}
}
\end{figure}

\beforepar\paragraph{Comparisons with a direct pixel loss.}
Fig.~\ref{fig:comploss} shows qualitative comparisons between the model trained with our full loss (Eq.~\ref{eq:total_loss}) and the same model trained with only an image loss, i.e., an $L_1$ loss between pixel values on the decoded image layer (with the decoder fixed).
The model trained with the image loss results in lower facial image quality and fewer facial variations. Our loss, measured at an early layer of the face decoder, allows for better supervision and leads to faster training and higher quality results.

%% file: sec5_conculsion.tex
\section{Conclusion}
\noindent
We have presented a novel study of face reconstruction directly from the audio recording of a person speaking. We address this problem by learning to align the feature space of speech with that of a pre-trained face decoder using millions of natural videos of people speaking. We have demonstrated that our method can predict plausible faces with the facial attributes consistent with those of real images. By reconstructing faces directly from this cross-modal feature space, we validate visually the existence of cross-modal biometric information postulated in previous studies~\cite{kim2018on,nagrani2018seeing}.
We believe that generating faces, as opposed to predicting specific attributes, may provide a more comprehensive view of voice-face correlations and can open up new research opportunities and applications.